\documentclass[a4paper,orivec]{llncs}

\usepackage{url}
\usepackage{amssymb}
\usepackage{amsmath} 
\usepackage[utf8]{inputenc}
\usepackage{xcolor}
\usepackage{multirow}
\usepackage{diagbox}
\usepackage{listings}     % Include the listings-package      
\usepackage{multirow}
\usepackage{lipsum}
\usepackage{arydshln}
\usepackage{slashbox}

\usepackage{graphicx}

\newcommand{\us}{\textsc{ObjCP-FP}}

\newcommand{\F}{\ensuremath{\mathbb{F}}}

\newcommand{\xf}[1]{\ensuremath{{#1}_{\F}}}
\newcommand{\itvf}[1]{\ensuremath{{\mathbf{#1}}_{\scriptscriptstyle\F}}}
\newcommand{\lbf}[1]{\ensuremath{\underline{#1}_{\scriptscriptstyle\F}}}
\newcommand{\ubf}[1]{\ensuremath{\overline{#1}_{\scriptscriptstyle\F}}}

\lstnewenvironment{code}[1][]
  {\lstset{
   numbers=left,
   xleftmargin=0.6cm,
     numberstyle=\footnotesize,
     escapeinside=@@,
     #1}}
      {}

\title{An efficient constraint based framework for handling floating point SMT problems}
\author{Heytem Zitoun\inst{1} \and Claude Michel\inst{2} \and Laurent Michel\inst{1} \and Michel Rueher\inst{2}}
\institute{University of Connecticut \\\email{ldm@uconn.edu}, 
\email{heytem.zitoun@uconn.edu}
\and
I3S (University Côte d'Azur - CNRS) \\\email{firstname.lastname@i3s.unice.fr}}

\begin{document}
\maketitle

\begin{abstract}
This paper introduces the 2019 version of \us{}, a novel Constraint Programming framework for floating point verification problems expressed with the SMT language of SMTLIB. SMT solvers decompose their task by delegating to specific theories (e.g., floating point, bit vectors, arrays, ...) the task to reason about combinatorial or otherwise complex constraints for which the SAT encoding would be cumbersome or ineffective. This decomposition and encoding processes lead to the obfuscation of the high-level constraints and a loss of information on the structure of the combinatorial model. In \us{}, constraints over the floats are first class objects, and the purpose is to expose and exploit structures of  floating point domains to enhance the search process. A symbolic phase rewrites each SMTLIB instance to elementary constraints, and eliminates  auxiliary variables whose presence is counterproductive. A diversification technique within the search steers it away from costly enumerations in unproductive areas of the search space. The empirical evaluation demonstrates that the 2019 version of \us{} is competitive on computationally challenging floating point benchmarks that induce significant search efforts even for other CP solvers. It highlights that the ability to harness both inference and search is critical. Indeed, it yields a factor 3 improvement over Colibri and is up to 10 times faster than SMT solvers. The evaluation was conducted over 214 benchmarks (The Griggio suite) which is a standard within SMTLIB.

\end{abstract}

\section{Introduction}
Embedded systems, and IoT devices in general, produce, analyze and consume significant volumes of data from arrays of sensors to carry out various control functions. Anti-lock breaking systems (ABS) offer 
a classic example of an embedded system that senses breaking actuation signals and wheel blockage to dynamically adapt the pressure on pads and prevent wheels from locking up and causing skidding. It is a cornerstone in modern automotive and the correctness of the control algorithm is paramount to the safety of vehicles. 
Such control systems play critical roles in sensitive devices in domains such as aerospace, transportation, health or even energy. Interestingly, the signals fed to the controllers are often represented as floating point numbers that are used in key calculations to decide how to actuate the device. 
It is common knowledge that \emph{programmers} often assume that floating point semantics (even IEEE~754~\cite{IEEE754}) are so similar to real numbers that one can use the former while assuming the semantics of the latter. It is, unfortunately, not the case and examples abound of erroneous controllers that caused significant losses~\cite{Bo,defense92}.

Program verification is undoubtedly a natural response to verify the correctness of controllers and ensure that calculations carried out over floating point values do not induce undesirable behaviors. Bounded model checking (BMC) (e.g.,~\cite{clarke2001}) was successfully used with a constraint programming framework to verify programs that used integer computations. Extending BMC to account for floating point is still a challenging problem.  

Satisfiability-Modulo-Theory (SMT) solvers are a natural extension of SAT solvers and are applied with success to the verification of floating point programs. Solvers such as Z3~\cite{z3}, MathSAT~\cite{mathsat5}, CVC4~\cite{cvc4} and SONOLAR~\cite{sonolar} are leading examples in this space. Fundamentally, SMT solvers \emph{decompose} their task by delegating to specific theories (e.g.,  arithmetic, arrays,  bit vectors, floating point,...) the task to reason about combinatorial or otherwise complex constraints for which the SAT encoding would be cumbersome or ineffective. The logical core of the problem remains in clausal form and is entrusted with a SAT solver that orchestrates the entire resolution. The resolution proceeds by doing inference and branching in the SAT solver (and the theory solvers) until a contradiction is detected. Upon reaching a contradiction, the responsible theory is tasked with producing a clause that precludes the repetition of decisions that yield the same contradiction. 
It is essential to realize that SMT solvers entrust the search responsibility entirely to the SAT solver and limit the responsibilities of the theories to produce inferences and learn clausal cuts. 
Some SMT solver may choose to `` bit blast'' some (or all) constraints to encode them in clausal form as well. It is undeniable that this decomposition and encoding processes lead to the obfuscation of the high-level constraints and a loss of information on the structure of the combinatorial model. It is also clear that decomposition and encoding can cause the proliferation of a large number of elementary constraints on the  bit level structures rather than the high-level representations. 
Such issues are critical when it comes to solving floating point verification problems that emphasize the use of complex nonlinear expressions.

Constraint Programming is an area of Artificial Intelligence with successes in a number of areas within combinatorial optimization. It is a de facto leading technology for scheduling and planning where commercial tools such as (Ilog~\cite{ILOG},CHIP~\cite{CHIP},...) are market leaders. Constraint Programming owes its successes to its ability to leverage strong inference capabilities within a flexible and programmable search framework. 
%The ability to derive strong inference is normally associated with global constraints such as allDifferent~\cite{regin94}, disjunctive~\cite{Carlier82} or cumulative~\cite{Aggoun93}. The global constraint catalog~\cite{Beldiceanu05} offers a list with more than 300 combinatorial structures for which potent constraints exists.
%
Search procedures dedicated to finite-domain CP solvers include weighted degree~\cite{Boussemart2004}, Impact-based search~\cite{Refalo2004}, Activity-based search~\cite{Michel2012} or counting search~\cite{Pesant2016} to name just a few. 
All those exploit the semantics of the application or the domain to make smart branching decisions. While distinctively different, floating point domains also offer unique properties that are exploitable to enhance the search process. Density~\cite{cp2017}, for instance, features a concentration of values that is not constant along the unit line and can be an indicator to focus (or not) on a variable.  
The promises of Constraint Programming are based on three prongs: 
First, semantic structures are preserved, exposed in the model, and exploitable by both the propagation engine and the search procedure.
Second, propagation algorithms tailored to specific structures offer stronger filtering and faster inferences.
Third, flexible search procedures can exploit the semantics of variables and the constraint network to prune doomed sub-trees and better branch on variables.

The crux of this paper is the composition of two key techniques to \emph{expose} and \emph{exploit} semantic structures. First, a symbolic phase is charged with rewriting each SMTLIB\footnote{For more information about the SMTLIB, see \url{http://smtlib.cs.uiowa.edu}.} instance to expose elementary constraints for which dedicated propagators exists. It also contributes to the elimination of auxiliary variables whose presence can negatively impact the search process.  
Second, a diversification technique within the search procedure steers it away from costly enumerations in unproductive areas of the search space. 
The empirical evaluation demonstrates that a Constraint Programming Solver is competitive on computationally challenging floating point benchmarks that induce significant search efforts even for other CP solvers. It highlights that the ability to harness both inference and search is critical. Indeed, \us{} yields a factor 3 improvement over Colibri \cite{colibri} (another CP solver) which uses a simple search procedure  and is up to $10$ times faster than SMT solvers. The evaluation was conducted over 214 benchmarks (The Griggio suite) which is a standard within SMTLIB. 

\paragraph{Contributions} 
This paper offers a novel constraint programming approach to solve floating point verification problems expressed with the SMT language of SMTLIB with the express purpose of exposing and exploiting structures. The implementation is compared to several state-of-the-art tools including the SMT solvers Z3, MathSAT, CVC4 and SONOLAR. It is also compared to Colibri, a CP solver with a strong inference engine designed specifically for floating point verification. %

\paragraph{}
The paper is organized as follows. Section~\ref{sec:cp} briefly recalls the foundation of constraint programming and specific details relevant in the context of floating point program verification. Section~\ref{sec:art} briefly recalls the state of the art in this space and positions CP technologies w.r.t. SMT solving. Section~\ref{sec:symb} discusses the symbolic reformulation techniques used to improve the encoding obtained from the SMTLIB specification and \emph{reveal} substructures present in the model whose semantics can be exploited within the propagation engine. Section~\ref{sec:search} reviews the heuristics (variable selection and domain splitting) as well as a key diversification strategy with a dramatic impact on performance. Section~\ref{sec:expe} discusses the empirical evaluation and section~\ref{sec:ccl} concludes the paper. 

\section{Constraint Programming}
\label{sec:cp}

Constraint programming derives its strength from the adoption of the mantra
\[
CP = Model + Search
\]
Indeed, CP solvers feature a declarative model dedicated to the specification of decision variables and the model constraints with a rich language featuring logical, arithmetic and combinatorial constraints. Unlike SAT~\cite{SAT} (or even MIP~\cite{MIP}) it is not limited to Boolean clauses or linear inequalities. Each constraint present in the model is supported by a dedicated filtering algorithm that exploits the semantics of the constraint to achieve strong filtering results with high efficiency.  
CP solvers also feature a powerful search component that is often used to author sophisticated search procedures that exploit specificities of the problem that are delicate to express in the model. While generic reusable searches exist, e.g.,~\cite{Refalo2004,Michel2012,Boussemart2004,Pesant2016}, examples of domain specific searches are commonplace, e.g.,~\cite{Tsang2003}. 

This section briefly reviews a few core concepts and their instantiation in the context of floating-point program verification. 

\begin{definition}[Constraint Satisfaction Problem or CSP]
A CSP is a triplet $\langle X,D,C \rangle$ where $X$ is a set of decision variables, $D$ is the Cartesian product of their domains, and $C$ is a set of constraints defined over subsets of X. The domain of a decision variable $x$ is denoted by $D(x)$. Note that $D = D(x_0)\times \cdots \times D(x_{n-1})$ when $X = \{x_0, \cdots , x_{n-1}\}$.
\end{definition}
\begin{definition}[Floating-Point Domain]
A domain $\itvf{D} = [\lbf{D}..\ubf{D}]$ denotes the set of floating points
$\{\xf{x} \in \F, \lbf{D} \leq \xf{x} \leq \ubf{D}\}$ with $\lbf{D} \in \F$ and $\ubf{D} \in \F$ where $\F$ is the set of floating point values. 
\end{definition}

\begin{definition}[Decision Variable]
A decision variable $x \in X$ is associated to a domain $D$ denoted $\itvf{D(x)}$ and it is instantiated if and only if $|\itvf{D(x)}| = 1$. 
\end{definition}

\begin{definition}[Constraint]
A constraint $c\in C$ of arity $n$ is a relation defined over a subset of $n$ variables from $X$ called its scope and denoted by $vars(c)$ (namely, $|vars(c)|=n$).
\end{definition}
Given a CSP $\langle X,D,C\rangle$, the task of a constraint solver is to find a domain $\itvf{D'}$ that delivers an instantiation of all variables. To do so, the solver engages in a search that alternates branching (by splitting the domain of a variable) with propagation to infer all the domain reductions that logically follow from the branching. Formally, at each node of such a search tree, the solver creates $2$ or more sub-problems with the addition of branching constraints and establishes 2B-consistency \cite{CDR99} within each sub-problem.
A satisfiable CSP $\langle X,D,C\rangle$ yields a domain $\itvf{D'} \subset \itvf{D}$ in which every variable $x \in X$ is instantiated in $\itvf{D'}$.
%and
%\[
%\bigwedge_{c \in C} c \mbox{ is 2B-consistent w.r.t. %} \itvf{D'} 
%\]
An unsatisfiable CSP $\langle X,D,C\rangle$ yields a refutation of the existence of an instantiated domain satisfying all constraints $c \in C$. 

Achieving 2B-consistency for a fixed constraint $c$ can be done with a dedicated algorithm that exploits the semantics of $c$. For instance, an elementary constraint such as $c \equiv x = y + z$ with $vars(c) =\{x,y,z\}$ uses \emph{projection functions} to compute new domains $\itvf{D'(x)}$, $\itvf{D'(y)}$ and $\itvf{D'(z)}$. Constraint programming solvers usually offer a rich collection of elementary constraints to express arithmetic relationships. It is essential to preserve such relationships when deriving a CSP from the program to be verified. Indeed, the introduction of auxiliary variables and the repetition of shared structure can weaken the strength of the propagation leading to the production of 2B-consistent domains that are \emph{bigger} than strictly necessary.

Adaptation of these principles to floating point arithmetic has its roots in the work of \cite{Claude2002} who introduced dedicated floating point projection functions.

%{ \color{blue}
%I guess we should add both following section right ?
\section{State of the Art}
\label{sec:art}
\paragraph{Satisfiability Modulo Theories (SMT).}
%The paper "Building Better Bit-Blasting for Floating-Point Problems" sent by Michel identify at least 2 class of SMT solver we are comparing with.
%\begin{itemize}
%    \item Default mathsat, z3 and SONOLAR use bit blasting techniques where floating-point constraints are directly converted into Boolean problems. The main disadvantage of those technique is to generate a huge Boolean formula (that can be mitigate by using approximation). 
%    \item techniques based on interval (abstract domain) : for instance ACDL framework implemented in mathsat (noted mathsatFP). This techniques allow SMT solver to get  tight domains (kind of filtering). 
%\end{itemize}
SMT solvers are widely used in program verification. They had great success especially in bounded model checking of integer programs.
SMT solvers can be seen as an extension of SAT solvers to decision procedures.
Indeed, modern SAT solvers are very efficient thanks to conflict clause learning.

Brillout et al~\cite{BrilloutKW09} introduced a bit blasting procedure to handle floating point verification problems in CBMC. It consists in modeling each floating point operation as a formula in propositional logic. The size of the formula is proportional to the number of operators and the width of the floating point type.
Well-known SMT solvers like Z3~\cite{ZeljicWR14}, MathSAT~\cite{BrainDGHK14} and CVC4~\cite{BrainSS19} followed suit, and included the same bit blasting techniques.

Yet, a direct application of the bit blasting procedure often leads to an explosion of the resulting formula size as a function of the number of floating operations and the size of the operand type. To circumvent this problem,  CBMC~\cite{BrilloutKW09} uses relaxations of the initial problem by means of under and over-approximations of each floating point operations. It generates a smaller propositional formula that may or may not be sufficient to solve the problem. As a consequence, many iterations with increasing precision and complexity might be required to determine the problem satisfiability.

As an alternative to bit blasting, MathSAT~\cite{BrainDGHK14}  also offers an interval propagation solver for floating point problems.  It relies on the abstract CDCL framework to allow conflict learning from the interval solver. Like our solver, it depends on a search strategy for the sake of completeness. However, the search strategy is quite simple: it chooses  variables in a round-robin order (lexical order) and splits the variable domain in two sub-intervals of the same cardinality.  Roughly speaking, this approach can be seen as a tight integration of a classical floating point constraint solver in a sat solver with the benefit of clause learning thanks the abstract CDCL framework.

\paragraph{Floating point constraint solvers.}
Floating point constraints introduced in~\cite{claude2001} filter
the domains of variables through interval arithmetic alone. 
FPCS (Floating Point Constraint Solver)~\cite{Claude2002} improved the capability to handle floating point constraints
by means of dedicated projection functions  that conform to floating point arithmetic.
Colibri~\cite{colibri} relies on the same principles to solve floating point CSP. 
However, it introduced distance constraints that strengthen the  variable domain reduction.
The 2017 version of the \us{} solver~\cite{cp2017} retained the projection functions of FPCS to enhance the underlying solver~\cite{objcp2013,objcp2017} with floating point constraints. This 2017 version also focused on search strategies to solve floating point constraint problems.

\paragraph{Contrasting CP and SMT.}
CP offers a high level of flexibility as well as abstractions to efficiently solve floating point verification problems.
CP searches are mainly based on variable domain exploration as opposed to the clause valuation strategies found in SMT solvers.
Critically, decisions found in floating point problems are based on heavier and more complex (non-linear) computations than what can be found in other programs.
A CP solver incrementally explores paths and decisions that determine a path with a higher and more precise knowledge of the computation states involved in these decisions. It thus has the potential to explore fewer paths to verify the program.

The path enumeration process found in SMT solvers build each path without having any information on the computations that affect the decisions involved along the path.
Though clause learning prunes the path enumeration by avoiding exploration of paths involving learned conflict, it does so at the expense of enumerating infeasible paths to identify these conflicts.
This requires solving condition affected by heavy computations and possibly, solving the same heavy computations repeatedly as they may lead to different paths.
The cost of such solving increases with the bit-blasting procedures used in solvers like Z3, CVC4 or MathSAT that produce propositional formula whose size grows with the number of operations and the precision. Even the MathSAT+ACDCL SMT solver, that combines a floating point interval solver with a SAT solver, is affected by this architecture.

A CP solver with an effective search strategy avoids the enumeration of many impractical paths as it maintains knowledge on all the computations and conditions that lead to the sub-tree of paths left to explore. Doing so, it avoids repeatedly solving costly subset of expensive constraints.

\section{Symbolic Reconstruction}
\label{sec:symb}
The SMTLIB language enables the specification of models in a solver neutral format. While the language is loosely based on LISP and allows for the specification of models that use multiple SMT theories, this paper focuses on the \texttt{QF\_FP} theory (Quantifier-Free Floating Point Theory). To illustrate, consider the \texttt{C} program computing a single step of the Newton method in Figure~\ref{fig:c-code}. 

\begin{figure}[t]
\begin{lstlisting}[language=C,numbers=none,frame=lines,numbers=left,
    stepnumber=1, basicstyle=\scriptsize]
float f(float x)  { return x - (x*x*x)/6.0f;}
float fp(float x) { return 1.0f - (x*x)/2.0f;}
int main() { 
  float r = x - f(x)/fp(x);
  assert(r < 0.1f);
  return 0;
}
\end{lstlisting}
\vspace{-2ex}
\caption{Example of source code for one step of the Newton method.}
\label{fig:c-code}
\end{figure}

This simple program can be \emph{encoded} as an SMTLIB text file (Figure~\ref{fig:realsmt}) that captures the sequence of low-level instructions visible in an SSA-style representation of the program. Such a translation can be done by a model checker like ESBMC~\cite{esbmc}. 
For instance, lines 5 and 6 declare $t_9$ and $t_{10}$ as two decision variables equal to the {\tt C} variables $x$ and $r$ at the initial state. Line 7 computes $t_9^2$ (i.e., $x^2$) using nearest rounding mode ($t_3$) and  stores it in $t_{13}$. Line 8 stores the constant 2 (expressed as a bit sequence) into $t_{12}$ while line 9 computes $\frac{x^2}{2}$ and stores it into $t_{13}$ before stating that $t_{14} = - t_{13}$ and adding the constant $1$ held in $t_{15}$ to deliver in $t_{16}$ the result of the function $fp$ in the {\tt C} code. Note how line 22 using a reified constraints to state $t_{25} = (t_{10} == t_{24})$ before forcing $t_{25}$ to be true on line 24. 

\begin{figure}[t]
\begin{lstlisting}[breaklines=true,numbers=left,
    stepnumber=1,frame=lines,basicstyle=\scriptsize]
(set-logic QF_FP)
(declare-fun x () (_ FloatingPoint 8 24))
(declare-fun r () (_ FloatingPoint 8 24))
(define-fun _t_3 () RoundingMode RNE)
(define-fun _t_9 () (_ FloatingPoint 8 24) x)
(define-fun _t_10 () (_ FloatingPoint 8 24) r)
(define-fun _t_11 () (_ FloatingPoint 8 24) (fp.mul _t_3 _t_9 _t_9))
(define-fun _t_12 () (_ FloatingPoint 8 24) (fp #b0 #b10000000 #b00000000000000000000000))
(define-fun _t_13 () (_ FloatingPoint 8 24) (fp.div _t_3 _t_11 _t_12))
(define-fun _t_14 () (_ FloatingPoint 8 24) (fp.neg _t_13))
(define-fun _t_15 () (_ FloatingPoint 8 24) (fp #b0 #b01111111 #b00000000000000000000000))
(define-fun _t_16 () (_ FloatingPoint 8 24) (fp.add _t_3 _t_15 _t_14))
(define-fun _t_17 () (_ FloatingPoint 8 24) (fp.mul _t_3 _t_11 _t_9))
(define-fun _t_18 () (_ FloatingPoint 8 24) (fp #b0 #b10000001 #b10000000000000000000000))
(define-fun _t_19 () (_ FloatingPoint 8 24) (fp.div _t_3 _t_17 _t_18))
(define-fun _t_20 () (_ FloatingPoint 8 24) (fp.neg _t_19))
(define-fun _t_21 () (_ FloatingPoint 8 24) (fp.add _t_3 _t_9 _t_20))
(define-fun _t_22 () (_ FloatingPoint 8 24)
(fp.div _t_3 _t_16 _t_21))
(define-fun _t_23 () (_ FloatingPoint 8 24) (fp.neg _t_22))
(define-fun _t_24 () (_ FloatingPoint 8 24) (fp.add _t_3 _t_9 _t_23))
(define-fun _t_25 () Bool (= _t_10 _t_24))
(assert _t_25)
(check-sat)
(exit)
\end{lstlisting}
\vspace{-5mm}
\caption{SMT encoding of the program in Figure~\ref{fig:c-code}.}
\label{fig:realsmt}
\end{figure}

\begin{figure}[t]
    \centering
    \includegraphics[scale=0.08]{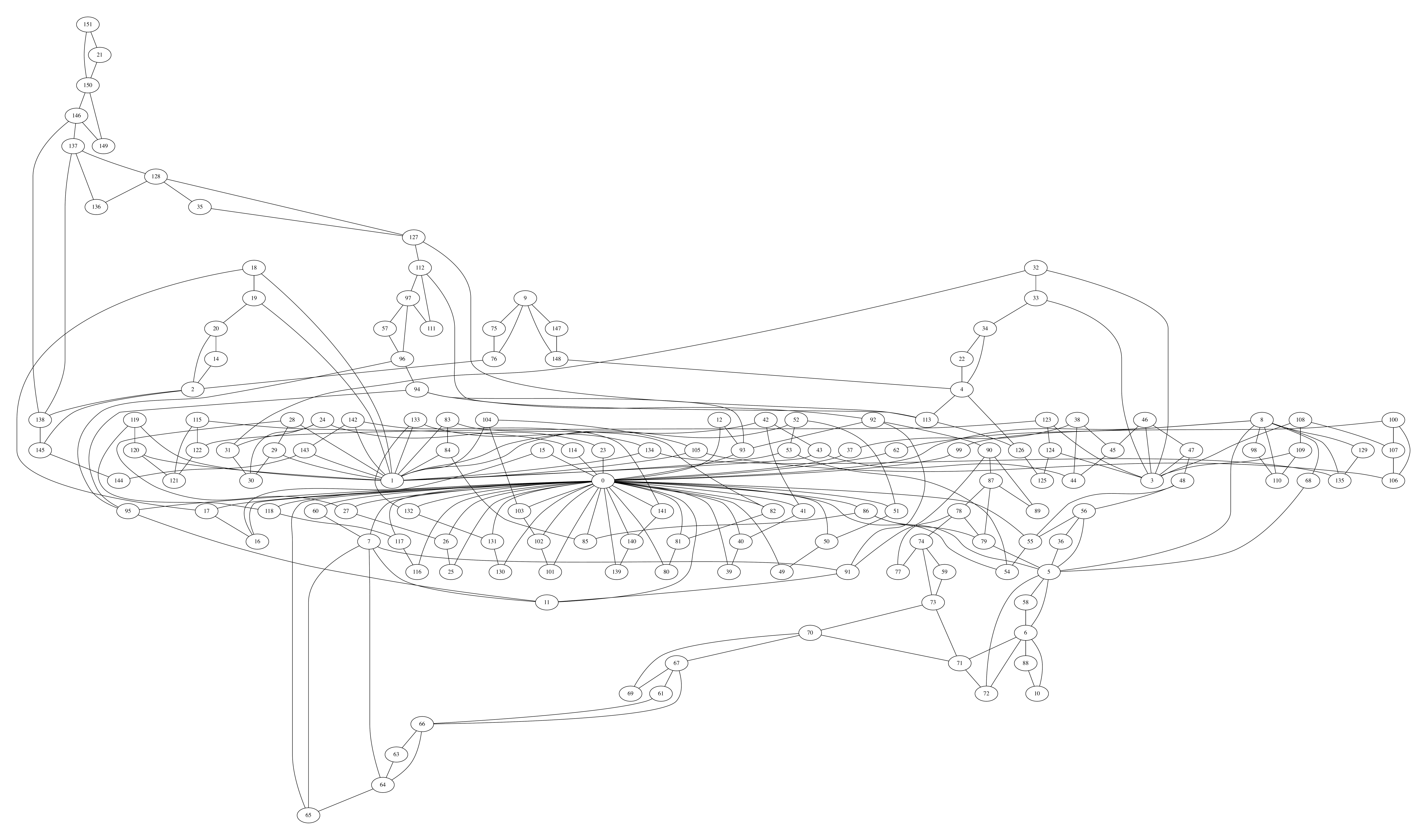}
    \vspace{-5mm}
    \caption{Constraints graph of {\tt sin2.c.2} for the default model (152 nodes and 242 edges)}
    \label{fig:def}
\end{figure}
\textbf{}
It is immediately apparent that the SMTLIB encoding decomposes every statement in the program with many auxiliary variables which may adversely impact intelligent search strategies that target the variables in the original {\tt C} program.  
Specifically and as an example, the {\tt sin2.c.2} benchmark yields a constraint graph in Figure~\ref{fig:def} with 152 nodes (decision variables) and 242 edges to convey the connectivity between variables. 

\begin{figure}[t]
    \centering
    \includegraphics[scale=0.08]{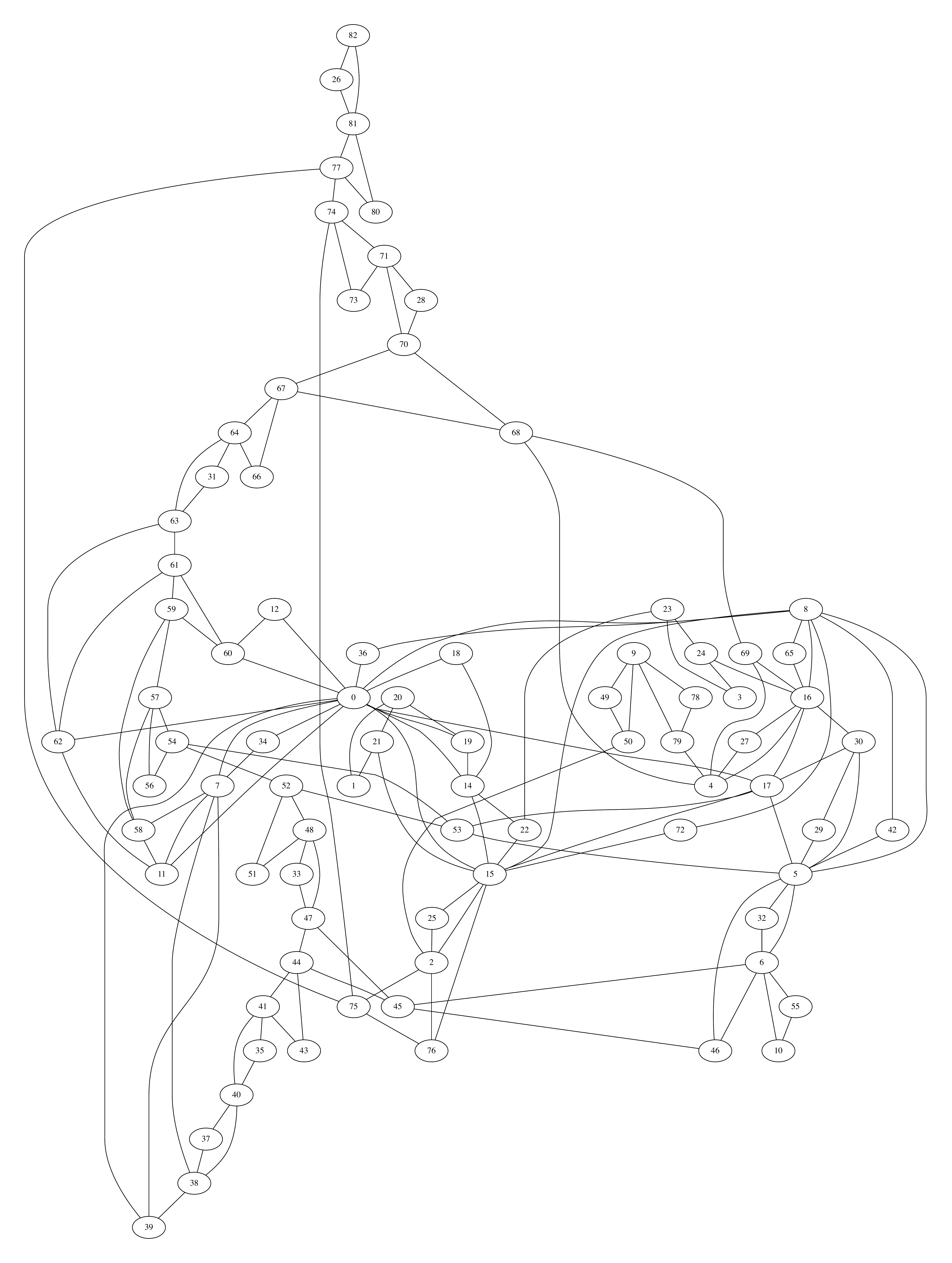}
    \vspace{-5mm}
    \caption{Constraints graph of {\tt sin2.c.2}  after CSE  (82 nodes and 147 edges)}
    \label{fig:cse}
\end{figure}

The purpose of this section is to outline the symbolic reconstruction that  simplifies this graph while preserving its semantics and \emph{exposing} key structures to enable the propagation engine of a CP solver to \emph{exploit} those structures. On the same {\tt sin2.c.2} benchmark, the reconstruction yields a graph that is \emph{half} the size of the original with only 82 nodes and 147 edges (see figure \ref{fig:cse}). 
Specifically, the section details the architecture of the reconstruction, its two phases and the impact it has on the filtering capabilities of the solver. 

% \begin{figure}
%     \centering
%     \begin{minipage}[t]{\textwidth}
%  %   \fbox{
%     \begin{minipage}[c]{0.55\textwidth}
%     \hline
%     \begin{minipage}[t]{.25\textwidth}
%         \begin{align*}
%         t_9 &= x\\
%         t_{10} &= r\\
%         t_{11} &= t_9 \times t_9\\
%         t_{12} &= 2.0\\
%         t_{13} &= \frac{t_{11}}{t_{12}}\\
%         t_{14} &= - t_{13}\\
%         t_{15} &= 1.0\\
%         t_{16} &= t_{15} + t_{14}\\
%         t_{17} &= t_{11} \times t_{9} \\
%         \end{align*}
%     \end{minipage}
%     \hfill
%     \begin{minipage}[t]{.25\textwidth}
%         \begin{align*}
%         t_{18} &= 6.0\\ 
%         t_{19} &= \frac{t_{17}}{t_{18}}\\
%         t_{20} &= - t_{19}\\
%         t_{21} &= t_9 + t_{20}\\
%         t_{22} &= \frac{t_{16}}{t_{21}}\\
%         t_{23} &= - t_{22}\\
%         t_{24} &= t_9 + t_{23}\\
%         t_{25} &= (t_{10} ==  t_{24}) \\
%         t_{25} &= 1
%         \end{align*}
%     \end{minipage}
%     \hline
%     \caption{SMT format encoding (prettified) of Figure~\ref{fig:c-code}}
%     \label{fig:smt-code}
%     \end{minipage}
% %    }
%     \hfill
% %    \fbox{
%     \begin{minipage}[t]{0.40\textwidth}
%     \hline
%     \begin{align*}
%         r &= x + \Bigg(-\cfrac{x + \Big(-\cfrac{x \times x \times x}{6.0}\ \Big)}{1.0+\Big(-\cfrac{x \times x}{2.0}\ \Big)}\Bigg)
%     \end{align*}
%     \hline
%     \caption{Example of abstract model}
%     \label{fig:R}
%     \end{minipage}
% %    }
%     \end{minipage}
% \end{figure}

\subsection{Architecture}
The purpose of the reconstruction is to first recover an \emph{abstract} model as close as possible to the original {\tt C} formulation while starting from the SMTLIB encoding.
Once an abstract model is available, it is easy to extract the set of decision variables appearing in the abstract model, and then to concretize it down so that it only uses elementary constraints. This final \emph{concrete} model is the basis for the resolution and the search procedure focuses on the variables identified in the abstract model. 

More formally, given a model $M_0$, a reconstruction function ${\cal R}$ and a concretization function ${\cal C}$, the objective is to reconstruct $M_1={\cal R}(M_0)$ to extract its decision variables  and then concretize $M_1$ into $M_2 = {\cal C}(M_1)$ to be solved by branching on $M_1$'s variables using $M_1$'s heuristic. Formally,

\[
\begin{array}{ccl}
     M_1 = \langle X_1,D_1,C_1\rangle = {\cal R}(M_0) & \rightarrow & \mbox{ yields } X_1  \\
     M_2 = \langle X_2,D_2,C_2\rangle = {\cal C}(M_1) & \rightarrow & \mbox{\tt solve}(M_2,X_1)
\end{array}
\]

\begin{figure}[t]
    \centering
    \begin{minipage}[t]{\textwidth}
    \begin{align*}
        r &= x + \Bigg(-\cfrac{x + \Big(-\cfrac{x \times x \times x}{6.0}\ \Big)}{1.0+\Big(-\cfrac{x \times x}{2.0}\ \Big)}\Bigg)
    \end{align*}
    \end{minipage}
    \caption{Reconstructed abstract model for {\tt Newton-1.1} via ${\cal R}$}
    \label{fig:R}
\end{figure}

\subsection{Reconstruction with ${\cal R}$}
\label{sec:reconstruction}

\noindent The reconstruction uses two steps that start from an initial model $M_0$. Those steps are identification and inlining. 

\paragraph{Identification} Partition the constraints in $M_0=\langle X,D,C\rangle$ into two disjoint sets $C_{aux}$ and $C_{main}$.  $C_{aux}$ is the set of  constraints defining auxiliary variables (i.e., constraints of the form $c \equiv x = \langle expr\rangle$ in which $\langle expr\rangle$ is an arbitrary expression and $x$ a variable) and $C_{main} = C \setminus C_{aux}$ is the remaining set of constraints. 

\begin{definition}[Auxiliary Variables]
Given a constraint $c \equiv x = \langle expr\rangle \in C_{aux}$, let $def(c)= \{x\}$ denote the set of auxiliary variables defined by $c$.
\end{definition}
Without loss of generality, let 
\[
def(C_{aux}) = \bigcup_{c \in C_{aux}} def(c)
\]
represents the lifted version of $def$ over a set of constraints. 

\paragraph{Inlining} Each constraint $c \in C_{main}$ is
rewritten by simply replacing all occurrences of auxiliary variables in $def(C_{aux})$ by their definitions. This substitution process ${\cal S}$ is applied until no further expansions can occur and the constraint $c'$ is irreducible under ${\cal S}$. Namely, it is the transitive closure ${\cal S}^*$ of ${\cal S}$. The final model drops all the auxiliary variables and their defining constraints and delivers CSP $M_1$:
\[
M_1 = \langle X \setminus def(C_{aux}) , 
        D' ,
        \bigcup_{c \in C_{main}} {\cal S}^*(c)
        \rangle
\]
in which $D'$ conveys domains for the variables in $C \setminus def(C_{aux})$. 
Clearly the inlining process shrinks the set of variables under consideration and eliminates all the constraints related to auxiliaries. This is the primary driver behind the reduction in model size. On the {\tt newton-1.1} example, it eliminates all but two variables ($x$ and $r$) and leaves only the single constraint\footnote{Note that the behavior of search heuristics can be (positively) impacted as the search may choose to focus on the variables identified by the reconstruction.} shown in Figure~\ref{fig:R}.

\subsection{Concretization with ${\cal C}$}
This phase leans on three steps and works from the model $M_1=\langle X_1,D_1,C_1\rangle$ produced by ${\cal R}(M_0)$. 

\paragraph{Factorization} 
The first step factors out common sub-expressions and  equates each one with a fresh variable injected where the common sub-expression used to be.  The composition of reconstruction and compilation can eliminate variables and simple equality constraints appearing in the SMT model such as $t_9$ and $t_{10}$  in Figure~\ref{fig:realsmt}.

\paragraph{Decomposition} Decompose each $c\in C_1$ into elementary constraints. The resulting model of this step is $M_2 = \langle X_2,D_2,C_2\rangle$, where variables $X_2 \supseteq X_1$ since $X_2$ contains all of $M_1$'s variables plus new auxiliaries produced by the factorization. 

\paragraph{Inequalities Cycles}
Occasionally, the {\tt C} source induces cycles of contradictory inequalities that make the model {\tt UNSAT}. Common contradictions are of the form: $\{X > Z, Y \leq Z, X=Y\}$. The pattern typically appear when a program has multiple conditional structure.   
While the solver can handle such cycles \emph{numerically}, this typically causes a slow convergence of the fixpoint algorithm executing at each search node. It is far better to recognize such a cycle symbolically and report the {\tt UNSAT} answer without further to do. 

\subsection{Impact of ${\cal C}({\cal R}(M_0))$}
Figure~\ref{fig:def} shows the constraint graph (graph where the nodes are variables and an edge between two variables exists if they are involved in the same constraint).  The corresponding constraints graph have 152 nodes and 242 edges (see figure \ref{fig:def}). 
Figure~\ref{fig:cse} uses the same benchmark and shows the constraint graph after the elimination of common sub-expressions from the abstract reconstructed model. This procedure drastically reduces the number of variables and constraints (reduction factor is 2 for this instance) leaving 82 nodes and 147 edges.

What is particularly important though, is that this reconstruction delivers better performance and a stronger filtering. For instance, the naive decomposition of equation~(\ref{eq:ex-square}):
\begin{equation}
    ((x+y) \times (x+y) = z)
    \label{eq:ex-square}
\end{equation}
introduces one auxiliary variable for each factor in the product, namely, two variables $a$ and $b$ and the constraint set:
\begin{equation}
a = x + y \land b = x + y \land z = a \times b
\end{equation}
Yet, the common sub-expression elimination introduces a single variable $a = x + y$ and a resulting decomposition:
\begin{equation}
a = (x + y) \land  z = a \times a   
\end{equation}
that a CP solver exploits since $z = a \times a$ is a square constraint for which a stronger filtering exists. 
These subtle effects on the inference capabilities of the solver are not negligible. Indeed, stronger filtering is valuable in its own right, but it also helps heuristics such as the variable selection discussed in the next section to differentiate between variables and make a better recommendation on whom to branch on next.

\section{Search}
\label{sec:search}
This section presents the search heuristic as well as the search strategies used in \us{}. In constraint programming search heuristics cover both the variable selection heuristics as well as the domain splitting heuristics\footnote{Also known as the value selection heuristics in finite domain solvers.}. 

\subsection{Variable Selection Heuristic}
The variable selection heuristic is based on a measure of density of variable domain introduced in~\cite{cp2017}. Roughly speaking, $dens(x)$, the density of variable $x$, is obtained by dividing the cardinality of its domain by its width.

Note that the domains of floating point numbers are not uniformly distributed, e.g., about half of the floats are in $[-1,1]$. Informally, \emph{dens} capture the proximity of floating point values in domains. Maximizing \emph{dens} captures variables with huge numbers of values in relation to the size of the domains. Such variables may have a larger number of values appearing in solutions.

\subsection{Domain Splitting Heuristic} 
When the search heuristic selects a variable, the solver must still divide the domain to explore the sub-problems. A 5-way split~\cite{valueHeuristic,cp2017} of a variable $x$ with domain $[L..U]$ creates five sub-problems (except in degeneracy case) where $D(x)$ is respectively restricted to 
$[L..L]$, $[U..U]$,$[M..M]$,$[L^+..M^-]$ and $[M^+..U^-]$, where $v^-$ and $v^+$ denote the floating point value immediately preceding (respectively, following) $v$. $M$  is the mid-point of the $[L..U]$ interval. In this implementation, the 5-way split is enhanced to pick $M$ differently and to allow the solver to focus on potentially interesting values by setting 
$M = middle(L,U)$ with the function $middle$ defined as
\[
middle(L,U) = \left\{
\begin{array}{ccl}
    0 & \Leftrightarrow & 0 \in [L..U] \\
    1 & \Leftrightarrow & 0 \notin [L..U] \wedge 1 \in [L..U] \\
    -1 & \Leftrightarrow & 0 \notin [L..U] \wedge 1 \notin [L..U] \wedge -1 \in [L..U] \\
    \frac{L}{2} + \frac{U}{2} & \Leftrightarrow& otherwise.
\end{array}\right.
\]
Interestingly, splitting on $1$ (or $-1$) has the added benefit that it can split a domain in equal-sized chunks. Indeed, about half the floating point numbers are in the range $[-1..1]$, so with a domain equal to $[0^+..+\infty]$, splitting at $1$ creates two sub-problems with equal-sized domains. 

Blending splitting and enumeration is particularly interesting. Notably, finite domain solvers often rely solely on enumeration while MIP and continuous domain solvers are notoriously relying on splitting alone. The combination of both techniques is valuable as enumerating on interesting values (either bounds or the split point) produces the smallest possible intervals for the variable branched on and can lead to a significant level of propagation that may not occur when only splitting. Yet, enumeration alone is impractical given the sheer domain size when looked at as a discrete set. 

\subsection{Diversification Strategy}
During the search, focusing on the same variable can be counter-productive and it may be wise to prevent the re-selection of the same variable at the next node. Focusing on the same variable at the top of the search tree can lead to a good reduction during propagation. But after a few re-selections, these reductions have less impact. Forcing the variable to change during the search leads to a more constrained problem and has a greater impact on filtering. 

Diversification techniques come from meta-heuristics. A classic example is Tabu search~\cite{Glover1989} which prevents the repetition of local moves that \emph{undo recent changes}. Tabu insists on shifting the focus to changes involving other variables, thereby driving the search in a different part of the 
search space. 
Tabu has rarely been considered in the context of a complete tree search\footnote{In~\cite{JL00}, the authors characterize a no-good technique for an hybrid algorithm that mixes global and local search as a Tabu mechanism. This is distinct from the use of Tabu here.}. Indeed, there is no risk to ``cycle'' and revisit the same configuration in a tree search and completeness guarantees that the whole space will be inspected, explicitly or implicitly. 
Yet, consider a variable with a large domain and a simple bisection heuristic for branching. The variable selection heuristic may find the variable appealing, and repeatedly branch on it until its domain density drops below others. Since density is not uniform\footnote{Note that while the interval $[0, 1]$ contains about a quarter of the floats while $[10^5, 10^5+1]$ contains only $128$ simple floats.}, and without a Tabu block, such obstinacy may trigger a dive to a large depth before switching to another variable. 

Formally, each time a variable $x$ is selected for branching at some tree node $n$ (at depth $depth(n)$), the search records a prohibition depth for $x$ equal to 
\[
last(x) = depth(n) + u
\]
Intuitively, the search is prohibited to consider $x$ again until the depth exceeds $last(x)$. When the search considers a  tree node $n$, the selection heuristic may instead choose to consider
\[
Y = \mathcal{S}(\{ y \in X | \neg bound(y) \wedge last(y) \leq depth(n) \}).
\]
As the value of the parameter $u$ increases, variables stay in ``purgatory'' increasingly longer after being selected. With $u=0$, this strategy simply reduces to normal branching and one must always have $u \leq |X|$ to operate\footnote{In practice, if $u$ exceeds the limit, it is set to the limit.}. 

\section{Experiments}
\label{sec:expe}
The following section evaluates \us{} and compares it to state-of-the-art solvers from the SMT community as well as Colibri, another solver from the CP community. The section discusses the benchmarks and reports on comparisons from multiple perspectives, wrapping up with an analysis of the relative merits of the two novel techniques introduced here. 

\subsection{Benchmarks}
The framework introduced in this paper is implemented in \us{} and the evaluation uses the {\tt QF\_FP}  SMTLIB benchmarks\footnote{\url{https://clc-gitlab.cs.uiowa.edu:2443/SMT-LIB-benchmarks/QF_FP}}. The evaluation focuses on the benchmark suite proposed by A.Griggio, i.e., a set of 214 benchmarks containing 117 satisfiable instances (i.e., program where a counter-example exists), and 97 unsatisfiable ones (i.e., correct programs). 
%Pour être précis, il y a 5 problèmes qui ne sont résolus par aucun solveur en 1mn. Parmis ces pbs, 3 sont unsat en moins d'une heure ... mais 2 restent non résolus (mul_03_3000_1 et test_v7_r7_vr1_c1_s24449) ... donc dans les 97 "UNSAT", il y en a quand même 2 dont on ne connait pas le status.

All the experiments are made on a Linux machine with an Intel Xeon CPU E5-2620 2.00 GHz and 16GB with a timeout of one minute. The comparison includes the CP solver  COLIBRI~\cite{colibri} and state-of-art SMT solvers : Mathsat~\cite{mathsat5}, Z3~\cite{z3}, CVC4~\cite{cvc4}, and SONOLAR~\cite{sonolar}. The comparison is based one the most recent (stable) version of each solver (i.e., COLIBRI v2176, Mathsat v5.5.4, Z3 v4.8.6, SONOLAR vDec2014, CVC4 v1.7).

In~\cite{cp2017}, one finds results on a collection of search heuristics for program verification that exploit properties of variables and their domain. It concludes that %\emph{absorption} and 
\emph{density} are particularly effective for variable selection heuristics. While the results included on this paper focus exclusively on the use of \emph{density}, \emph{all search heuristics} benefit from the techniques described in this paper. 

\subsection{Reporting and Validation}
In the following tables, the time needed for symbolic processing is included in the solving time. All run times are in minutes except Table~\ref{tab:vbs} which uses seconds. 

Note that all results were cross-checked for correctness against SMT solvers through the injection of the solutions  (in case of SAT instances) for a pure instantiation check. 
The conjunction of the solution and the initial model is given to other solvers to check that the resulting model is indeed correct. For {\tt UNSAT} benchmarks, we check that the result is consistent with SMT solvers (also {\tt UNSAT}). 

\subsection{Results and Discussion}
% {\color{blue} 
%  Mentionner que les résultats de notre solveur ont été vérifié et comment ils l'ont été. 
% 
%  Note that the results produced by oour solver have been checked against other solvers using the following procedure: for each problem that is satisfiable, we have check that the conjunction of the produced solution and the initial problem is still a satisfiable problem using other solvers while for unsatisfiable problems, we checked that there is no discrepancy between this result and results produced by other solvers on the same problem.
%  (à améliorer)
% 
%
%  Donner le nombre de problèmes pour la table 2 (préciser le nombre de SAT et UNSAT))
%  Supprimer VBS de l'intitulé des colonnes de la table 3 (sauf dernier .. ou mettre All)
%  Mettre CP avant Other dans table 3.
%  Titre des tables 2 et 3.
% 
% }
The main results of the paper are reported in Table~\ref{tab:full} which compares all solvers with the proposed approach (i.e., \us{}). The table reports the percentage of solved instances, the number of timeout and the total time (in minutes). The results are broken down into two categories for benchmarks with solutions ({\tt SAT}) and benchmarks without solution ({\tt UNSAT}). A last set of rows reports aggregate results ({\tt ALL}). 

\begin{table}[t]
\centering
 \begin{tabular}{|l l | r r r r r r| r |} 
 \hline
\multicolumn{2}{|c|}{Solver} & \multicolumn{2}{c}{Mathsat}  & \multirow{2}{*}{Z3} & \multirow{2}{*}{CVC4} & \multirow{2}{*}{Sonolar} & \multirow{2}{*}{Colibri} & \multirow{2}{*}{\us}\\ 
& & Default & ACDCL & & & &  & \\ 
 \hline\hline
 \multirow{3}{*}{{\tt SAT}}
 & \% solved & 58.97 & 64.1  & 29.06 & 65.81 & 70.09 & 82.05 & \textbf{99.15}\\ 
 & TO & 48 & 42 & 83 & 40 & 35 & 21 & \textbf{1}\\
 & Time (m) & 65.5 & 45.38 & 88.71 & 54.62 & 48.4 & 28.08 & \textbf{3.12} \\
 \hline
 \multirow{3}{*}{{\tt UNSAT}} & \% solved & 54.64 & 74.23 &  20.62 & 52.58 & 60.82 & 72.16 &  \textbf{88.66} \\
 & TO & 44 & 25 & 77  & 46  & 38 & 27 & \textbf{11}\\
 & Time (m) & 58.17 & 27.57 & 84.34 & 54.9 & 44.65 & 33.7 & \textbf{14.05} \\
 \hline
\multirow{3}{*}{{\tt ALL}} 
 & \% solved & 57 & 68.69 & 25.23 & 60 & 65.89 & 77.57 &  \textbf{94.86} \\
 & TO & 92 &  67 & 160 & 86 & 73 & 48 & \textbf{12} \\
 & Time (m) & 124.68 & 72.95 & 173.05 & 109.05 &  94.05 &  61.9  &  {\bf 17.18} \\
 \hline
\end{tabular}
\vspace{1ex}
\caption{Comparison of the different approach}
\label{tab:full}
\vspace{-10mm}
\end{table}

While {\tt MathSAT-ACDCL} is undeniably the strongest contender among SMT solvers, it is dominated by Colibri which is itself outperformed in all categories by \us{}. Indeed, \us{} leads in the fraction of solved instances and in the CPU time spent in doing so. The margin is also significant as \us{} is 4.2 times faster than {\tt MathSAT-ACDCL} and produces 5.6 times fewer timeouts. The dominance is present in both SAT and UNSAT categories but is particularly striking on SAT instances where the best SMT solver ({\tt SONOLAR}) produces 35 timeouts against only \emph{one} for \us{}. Almost all SAT benchmarks except one are actually solved by our solver in less than 4 minutes. The second best solver, i.e., Colibri fails on 21 of these benchmarks and requires almost half an hour of CPU. 
When considering the aggregate results, it becomes quite evident that solvers that exploits the problem semantics with dedicated filtering (through a theory as {\tt MathSAT-ACDCL} or via CP -- Colibri and \us{}--) lead the pack. Yet, a dedicated search and an even sharper focus on exploiting structures paid off handsomely for \us{}. 
A second observation can be made when contrasting the results of each solver in the SAT and UNSAT categories. All solvers except {\tt MathSAT-ACDCL} do much better on SAT instances than on UNSAT instances and \us{} fits that description too (with 99.15\% of SAT instances solved and 88.66\% of UNSAT instances solved). 
{\tt MathSAT-ACDCL} appears to favor UNSAT instances with 74\% solved UNSAT vs. 64\% solved SAT. 

\paragraph{Comparison on Tractable Instances}
Table~\ref{tab:noto} focuses on the 44 benchmarks solved by every solver. This set of benchmarks includes 30 satisfiable and 14 unsatisfiable instances. The recorded times are in seconds. Even when focusing on the set of problems solved by everyone, \us{} is still substantially faster (from 4 times faster to 60 times).

\begin{table}[t]
\centering
 \begin{tabular}{|l| r c r r r r| r |} 
 \hline
\multirow{2}{*}{\backslashbox{Time (sec)}{Solver}} & \multicolumn{2}{c}{Mathsat}  & \multirow{2}{*}{Z3} & \multirow{2}{*}{CVC4} & \multirow{2}{*}{Sonolar} & \multirow{2}{*}{Colibri} & \multirow{2}{*}{\us}\\ 
& Default & ACDCL & & & &  & \\ 
 \hline\hline
 {\tt SAT} & 143.1 & 19.3 & 216.1 & 48.8 & 23.7 & 19.6 & \textbf{4.6} \\
 \hline
 {\tt UNSAT} & 57.9 & 22.8 & 216.7 & 56.8 & 19.1 & 9.6 & \textbf{2.5} \\
 \hline
{\tt ALL}  & 201 &  42.1 & 432.8 & 105.6 &  42.8 &  29.2  &  {\bf 7.1} \\
 \hline
\end{tabular}
\vspace{1ex}
\caption{Comparison of all solvers over the subset of benchmarks solved by every tool}
\label{tab:noto}
\end{table}

\paragraph{Virtual Best Solver Comparison}
Table~\ref{tab:vbs} compares 5 different Virtual Best Solver (VBS). These virtual solvers corresponds to ``solver portfolios'', i.e., for each benchmark, the result of the VBS is the result of the best solver for that particular benchmark.  Five VBS are reported: 
\begin{itemize}
\item {\tt BBSMT} is the bit-blasting VBS $\{$MathSAT-Default,Z3,CVC4,Sonolar$\}$. \item {\tt ALLSMT} is  {\tt BBSMT} $\cup$ $\{$Mathsat-ACDCL$\}$.
\item {\tt CP} is $\{$ Colibri , \us{} $\}$.
\item {\tt OTHER} is {\tt ALLSMT} $\cup$ $\{$Colibri$\}$.
\item {\tt ALL} is the VBS which includes all solvers. 
\end{itemize}

\begin{table}[t]
\vspace{-2mm}
\centering
 \begin{tabular}{|l l | r r r r | r |} 
 \hline
 \multicolumn{2}{|c|}{Solver} & BBSMT & ALLSMT &  CP & OTHER  & ALL \\ 
 \hline\hline
 \multirow{3}{*}{{\tt SAT}}
 & \% solved & 81.2 & 82.05 & 100 & 94.02 & 100 \\
 & TO & 22 & 21 & 0 & 7 & 0\\
 & Time (m) &  41.06 & 32.41 & 2.13 & 17.28 & 2.06  \\
 \hline
 \multirow{3}{*}{{\tt UNSAT}} & \% solved & 62.89 &  88.66 & 93.81 &  88.66 & 94.85 \\
 & TO & 36 & 11 & 6 & 11 & 5\\
 & Time (m) & 42.61 & 14.74 & 7.92 & 13.62 & 7.64\\
 \hline
\multirow{3}{*}{{\tt ALL}} 
 & \% solved & 72.9 & 85.05 & 97.2 & 91.59 & 97.66\\
 & TO & 58 & 32 & 6 & 18 & 5 \\
 & Time (m) & 83.67 & 47.15 & 10.05 & 30.89  & 9.7 \\
 \hline
\end{tabular}
\vspace{1ex}
\caption{Comparison of 5 VBS}
\label{tab:vbs}
\vspace{-10mm}
\end{table}

Naturally, the ALLSMT VBS that includes a solver using the semantics of constraints dominates the pure bit-blasting approaches across the board. The gains are particularly telling in the UNSAT category. Second, the CP VBS, which includes only two solvers, delivers only 6 timeouts and wraps up the entire computation in under 10 minutes or nearly 5 times faster than the SMT VBS. Finally, observe how adding the SMT VBS to the CP VBS (i.e., ALL) barely improves over a CP approach. 
\emph{Interestingly the results from every solver but \us{} fail to improve over \us{}}. 

% \begin{itemize}
%     \item VBS-CP is virtually the same than ALL
%     \item OTHER is worst than \us{} (reverse ...)
%     \item 
% \end{itemize}

\paragraph{Contributions of each technique}
Table~\ref{tab:options} presents the impact of each technique used by our solver. The column {\tt NONE} corresponds to the 2017 version of \us{} without any new techniques. {\tt DIV} is \us{} augmented by diversification. {\tt CSE}'s column is \us{} with common sub-expression elimination. Column {\tt DIV+CSE} is \us{} with both techniques included. Finally, {\tt DIV+CSE+CY}, which corresponds to the 2019 version of \us{},  adds cycle elimination to {\tt DIV+CSE}. The table records the percentage of solved instances, the number of timeouts and the total time (in minutes) for each configuration ({\tt SAT}, {\tt UNSAT} and {\tt ALL}).

\begin{table}[t]
\centering
\begin{tabular}{|l|l|r|r|r|r|r|}
\hline
 type &    measure & {\tt NONE} & {\tt DIV} & {\tt CSE} & {\tt DIV+CSE} & {\tt DIV+CSE+CY}\\
\hline
\multirow{3}{*}{{ \tt SAT}} &  \% Solved &  77.12 &  83.9  &  88.14 &  99.15&  99.15\\
     &         TO &     27 &     19  & 14 & 1& 1 \\
    &       T(m) &  30.14 &  22.94 &  14.36 & 3.12& 3.12\\
\hline
\multirow{3}{*}{{\tt UNSAT}} &  \% Solved &  59.79 &  84.54 & 69.07 & 87.63 & 88.66\\
        &         TO &     39 &     15 & 30 &     12 & 11\\
        &       T(m) &  44.03 &   21.2 &  35.85 &  15.05 & 14.05 \\
\hline
\multirow{3}{*}{{\tt ALL}} &   \% Solved &  69.16 &  84.11 &  79.44&  93.93 &  94.39\\
      &         TO &     66 &     34 &     44&     13&     12\\
      &       T(m) &  74.17 &  44.14&  50.21&  18.18 &  17.18\\
        \hline
\end{tabular}
\vspace{1ex}
\caption{Comparison of different options for \us}
\label{tab:options}
\vspace{-8mm}
\end{table}

CY allows to solve only one more problem.
More interestingly, {\tt DIV} and {\tt CSE} are two effective and additive (or near additive) techniques and thus fully orthogonal. Namely, they contribute to improvements to (mostly) disjoints sets of benchmarks indicating a huge gain for doing both as shown in column {\tt DIV+CSE}. Perhaps counter-intuitively, {\tt DIV} is quite effective on UNSAT benchmarks where the percentage of solved instances jumps from 60\% to 84.6\% while the gains on SAT instances attributable to {\tt DIV} are far more modest (+6\%). 
Augmenting \us{} by diversification technique increases the percentage of solving benchmarks by 15\% and reduces the required time by 30 minutes. Likewise, augmenting \us{} with common sub-expression elimination increases this percentage by 10\% and reduces the total time by 24 minutes. When both are combined, 25\% (15\% of diversification + 10\% of common sub-expression elimination) new benchmarks are now solved. The solving time is reduced by 56 minutes : more than 30 gained by diversification and 24 gained by CSE.

\section{Conclusion}
\label{sec:ccl}

In this paper we have introduced the 2019 version of \us{}, a novel and efficient constraint programming framework for floating point verification problems expressed with the SMT language of SMTLIB. In this framework, constraints over the floats are first class objects, and thus \us{} can take advantage of the structures of floating point domains to boost the search process. The symbolic rewriting step that eliminates useless auxiliary variables and the diversification techniques implemented within the search are two key features of this new solver. The experiments demonstrate that the abstraction level and the flexibility of constraint programming are very well adapted for solving standard floating point benchmarks of the SMT community.  A detailed analysis of the experiments shows that \us{} is particularly effective on difficult problems that require significant numerical calculations, whereas SAT based solver are better on problems that require limited numerical computations. Future work could focus on a deeper analysis of the specific capabilities of the different types of solvers, and on a more accurate analysis of the complexity of the required numerical computations.

\bibliographystyle{plain}
\bibliography{tacas.bib}

\end{document}